\pdfoutput=1

\documentclass[11pt]{article}

\usepackage{EACL2023}
\usepackage{times}
\usepackage{latexsym}
\usepackage{graphicx} 
\usepackage{bm} 
\usepackage{booktabs}

\usepackage[T1]{fontenc}

\usepackage[utf8]{inputenc}

\usepackage{microtype}

\usepackage{inconsolata}

\title{Translate First Reorder Later: \\ Leveraging Monotonicity in Semantic Parsing}
\author{Francesco Cazzaro\footnotemark[1] \and Davide Locatelli\Thanks{ Equal contribution}\footnotemark[1] \and Ariadna Quattoni \\
  Universitat Politècnica de Catalunya \\
  \texttt{name.lastname@upc.edu} \\ \\ \AND
  Xavier Carreras \\
  dMetrics \\ 
  \texttt{xavier.carreras@dmetrics.com}
  }

\begin{document}
\maketitle
\begin{abstract}

Prior work in semantic parsing has shown that conventional seq2seq models fail at compositional generalization tasks. This limitation led to a resurgence of methods that model alignments between sentences and their corresponding meaning representations, either implicitly through latent variables or explicitly by taking advantage of alignment annotations. We take the second direction and propose \textsc{TPol}, a two-step approach that first translates input sentences monotonically and then reorders them to obtain the correct output. This is achieved with a modular framework comprising a \emph{Translator} and a \emph{Reorderer} component. We test our approach on two popular semantic parsing datasets. Our experiments show that by means of the monotonic translations, \textsc{TPol} can learn reliable lexico-logical patterns from aligned data, significantly improving compositional generalization both over conventional seq2seq models, as well as over other approaches that exploit gold alignments. Our code is publicly available at \url{https://github.com/interact-erc/TPol.git}

\end{abstract}

\section{Introduction}
\label{intro}
The goal of a semantic parser is to map natural language sentences (NLs) into meaning representations (MRs). Most current semantic parsers are based on deep sequence-to-sequence (seq2seq) approaches and presume that it is unnecessary to model token alignments between NLs and MRs because the attention mechanism can automatically learn the correspondences \citep{dong-lapata-2016-language, jia-liang-2016-data}. However, recent work has shown that such seq2seq models find compositional generalization challenging, i.e., they struggle to predict unseen structures made up of components observed at training \citep{pmlr-v80-lake18a, finegan-dollak-etal-2018-improving}. 

This limitation motivated the resurgence of approaches that model alignments between NL sentences and their corresponding MRs more similarly to classical grammar and translation-based parsers \citep{herzig-berant-2021-span}. Alignments can be modeled either implicitly through latent variables \citep{wang-etal-2021-structured-nips}, or explicitly by leveraging gold alignment annotations \citep{shi-etal-2020-potential, liu-etal-2021-learning-algebraic}. We take the second direction and exploit a recently released multilingual dataset for semantic parsing annotated with word alignments: \textsc{GeoAligned} \citep{locatelli-quattoni-2022-measuring}, which augments the popular \textsc{Geo} benchmark \citep{aaai-zelle-1996}. 

\begin{figure}
    \centering
    \includegraphics[width=\linewidth]{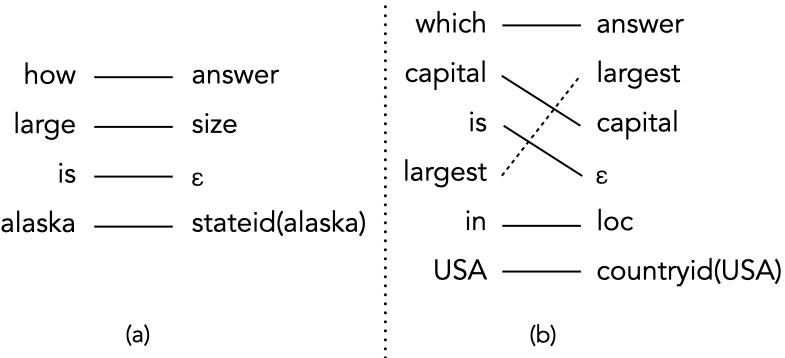}
    \caption{Examples from the \textsc{GeoAligned} dataset. (a) is a monotonic alignment, (b) is non-monotonic.}
    \label{fig:example-alignments}
\end{figure}

Figure \ref{fig:example-alignments} shows some examples of the annotations provided. One key observation is that a significant percentage of the alignments are monotonic, i.e., they require no reordering of the target MR (Figure \ref{fig:example-alignments}a), as opposed to non-monotonic alignments (Figure \ref{fig:example-alignments}b). This suggests that learning reliable lexico-logical translation patterns from aligned data should be possible. If there are simple patterns, shouldn't an ideal model be able to exploit them? 

With this in mind, we propose \textsc{TPol}, a \underline{T}wo-step \underline{P}arsing approach that leverages m\underline{o}notonic trans\underline{l}ations. \textsc{TPol} introduces a modular framework with two components: a \emph{Monotonic Translator} and a \emph{Reorderer}. The Translator is trained from pairs of NLs and MRs, where the MRs have been permuted to be monotonically aligned. Hence, the Translator's output will be an MR whose order might not correspond to that of the gold truth. For this reason, the Reorderer is trained to restore the correct order of the original MR.

Our experiments on \textsc{GeoAligned} demonstrate that compared to a multilingual BART model \citep{liu-etal-2020-multilingual-denoising}, \textsc{TPol} achieves similar performance on the random test split but significantly outperforms on the compositional split across all languages. For example, on the query split in English, mBART obtains $69.4\%$ in exact-match accuracy and \textsc{TPol} obtains $87.8\%$. This result also improves on the $74.6\%$ obtained by \textsc{SpanBased} \citep{herzig-berant-2021-span}, another approach that leverages alignment annotations.

Because most semantic parsing datasets do not contain alignment information, we experiment with alignments generated automatically. On \textsc{Geo}, \textsc{TPol} trained with automatic alignments still outperforms mBART, and in particular on the English query split it improves by almost 10 points. Furthermore, we show competitive results on the popular \textsc{Scan} dataset \citep{pmlr-v80-lake18a}.

In summary, the main contributions of this paper are: 
\begin{enumerate}
    \item We propose \textsc{TPol}, a modular two-step approach for semantic parsing which explicitly leverages monotonic alignments;
    \item Our experiments show that \textsc{TPol} improves compositional generalization without compromising overall performance;
    \item We show that even without gold alignments \textsc{TPol} can achieve competitive results.
\end{enumerate}

\section{Related Work}
\label{related_work}

Recently, the semantic parsing community has raised the question of whether current models can generalize compositionally, along with an effort to test for it \citep{pmlr-v80-lake18a, finegan-dollak-etal-2018-improving, kim-linzen-2020-cogs}. The consensus is that conventional seq2seq models struggle to generalize compositionally \citep{loula-etal-2018-rearranging, keysers2020measuring}. Moreover, large pre-trained language models have been shown not to improve compositional generalization \citep{oren-etal-2020-improving, qiu-etal-2022-evaluating}. This has prompted the community to realize that parsers should be designed intentionally with compositionality in mind \citep{ lake-2019-compositional, gordon-etal-2020-permutation, weissenhorn-etal-2022-compositional}. It has also been pointed out that compositional architectures are often designed for synthetic datasets and that compositionality on non-synthetic data is under-tested \citep{shaw-etal-2021-compositional}.

Data augmentation techniques have been proposed to improve compositional generalization \citep{andreas-2020-good, yang-etal-2022-subs, qiu-etal-2022-improving}. Another strategy is to exploit some level of word alignments. In general, there has been a resurgent interest in alignments as it has been shown that they can be beneficial to neural models \citep{shi-etal-2020-potential}. It has also been conjectured that the lack of alignment information might hamper progress in semantic parsing \citep{zhang-etal-2019-amr}. As a result, the field has seen some annotation efforts in this regard \citep{shi-etal-2020-potential, herzig-berant-2021-span, locatelli-quattoni-2022-measuring}. 

Alignments have been modeled implicitly: \citet{wang-etal-2021-structured-nips} treat alignments as discrete structured latent variables within a neural seq2seq model, employing a framework that first reorders the NL and then decodes the MR. Explicit use of alignment information has also been explored: \citet{herzig-berant-2021-span} use alignments and predict a span tree over the NL. \citet{sun-etal-2022-leveraging} recently proposed an approach to data augmentation via sub-tree substitutions. In text-to-SQL, attention-based models that try to capture alignments have been proposed \citep{lei-etal-2020-examining, liu-etal-2021-awakening}, as well as attempts that try to leverage them directly \citep{sun-etal-2022-leveraging}.

Our two-step approach resembles statistical machine translation, which decomposes the translation task into lexical translation and reordering \citep{chang-etal-2022-anticipation}. Machine translation techniques have previously been applied to semantic parsing. The first attempt was by \citet{wong-mooney-2006-learning}, who argued that a parsing model can be viewed as a syntax-based translation model and used a statistical word alignment algorithm. Later a machine translation approach was used on the \textsc{Geo} dataset, obtaining what was at the time state-of-the-art results \citep{andreas-etal-2013-semantic}. More recently, \citet{agarwal-etal-2020-machine} employed machine translation to aid semantic parsing.

\section{Preliminaries: Word Alignments}
This section briefly explains word alignments, showing the difference between monotonic and non-monotonic alignments, and illustrates the notion of monotonic translations. 

Assume that we have a pair of sequences ${\bf x} = x_1, \ldots, x_{n}$ and ${\bf y} = y_1, \ldots, y_{m}$, where $n$ and $m$ are the respective sequence lengths. A bi-sequence is defined as the tuple $({\bf x}, {\bf y})$. In our application, ${\bf x}$ is a NL sentence, and ${\bf y }$ is its corresponding MR. For example:\\

\noindent ${\bf x}=$ which city has the highest population density? 

\noindent ${\bf y}=$ \texttt{answer(largest(density(city(all))))}\\

A word alignment is a set of bi-symbols $\mathcal{A}$, where each bi-symbol defines an alignment from a token in the NL to a token in the MR. For instance, the bi-symbol $(x_i,y_j)$ aligns token $x_i$ to token $y_j$. In our example, the tokens "which" and "answer" could be paired by a bi-symbol $($which, answer$)$.

If a token $x_i$ does not align to anything in ${\bf y}$, an $\varepsilon$ is introduced in ${\bf y}$: the resulting bi-symbol $(x_i, \varepsilon)$ corresponds to a deletion. In our example, the token "has" in the NL can be deleted with a bi-symbol $($has, $\varepsilon)$. Similarly, if a token $y_j$ is not aligned to a token in ${\bf x}$, an $\varepsilon$ is introduced in ${\bf x}$: $(\varepsilon, y_j)$ is an insertion. In our example, the token "all" in the MR is inserted with bi-symbol $(\varepsilon$, all$)$.

The bi-symbols in $\mathcal{A}$ are all one-to-one. Hence, to map a single token to a phrase, i.e., to multiple tokens, it is necessary to choose a head token in the phrase, while the remaining tokens require insertion or deletion. In our example, the token "density" in the MR corresponds to "population density" in the NL, and, if "density" is chosen as the head token in the NL, "population" needs a deletion: the alignment will be given by the bi-symbols $($population, $\varepsilon)$ and $($density, density$)$.\footnote{\citet{locatelli-quattoni-2022-measuring} showed that annotators are consistent in the way they pick head-tokens, and reported high inter-annotator agreement scores on \textsc{GeoAligned}.} Following this strategy, this notation can account for one-to-many and many-to-one alignments with deletion and insertion operations.

\begin{figure}
    \centering
    \includegraphics[width=\linewidth]{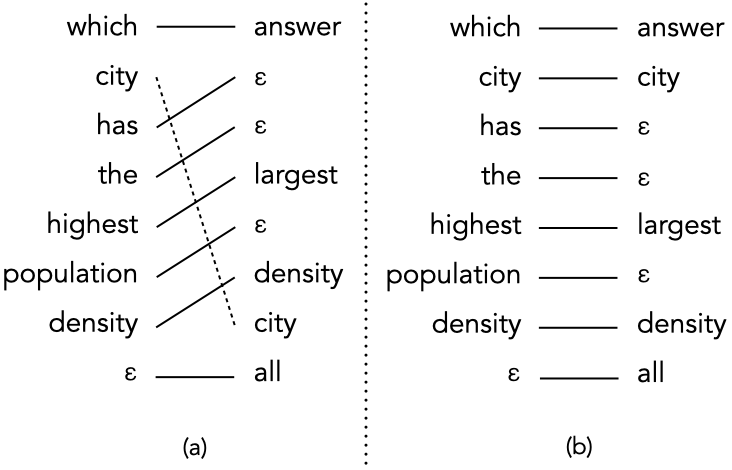}
    \caption{(a) A possible alignment for an NL-MR pair. (b) The corresponding monotonic translation. For simplicity, we removed the brackets and question mark.}
    \label{fig:example-alignment-definition}
\end{figure}

Figure \ref{fig:example-alignment-definition}a shows a possible bi-sequence word alignment for the aforementioned example. Each bi-symbol is conveniently represented by a horizontal line connecting the tokens it aligns. 

Alignments can be monotonic or non-monotonic. An alignment is monotonic if it does not involve any \emph{crossing}, i.e., a mapping that does not require reordering tokens. In our example, the alignment is non-monotonic because the bi-symbol $($city,city$)$ crosses over others. By permuting the MR, we can obtain a monotonic translation of the NL: Figure \ref{fig:example-alignment-definition}b shows such permutation. The next section illustrates how \textsc{TPol} can leverage these translations.

\section{Translate First Reorder Later}
\label{technical_section}

\begin{figure*}
    \centering
    \includegraphics[width=\linewidth]{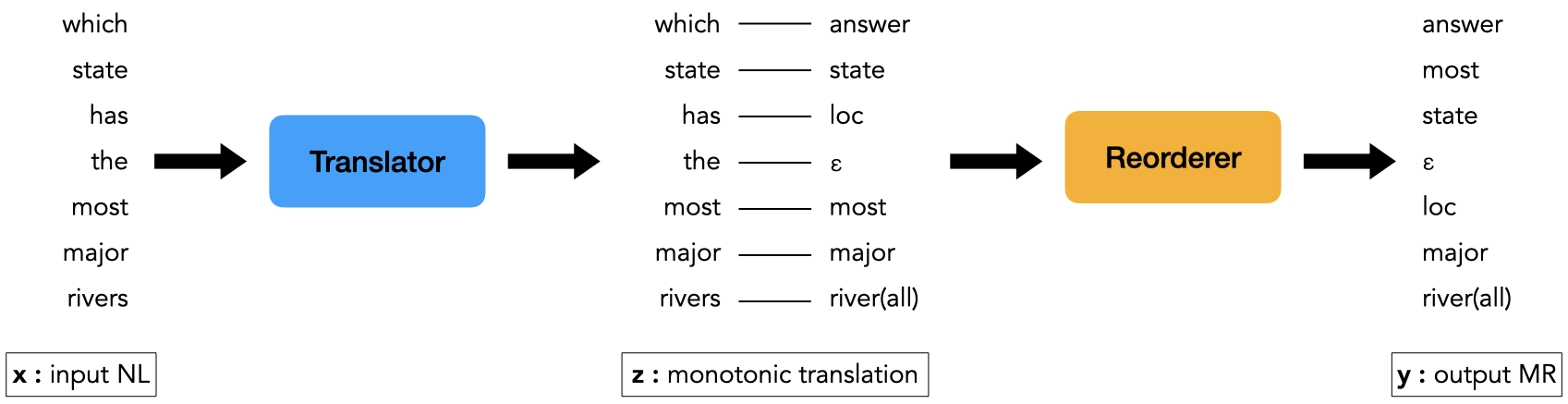}
    \caption{The \textsc{TPol} parsing approach. An input sentence $\bf x$ is fed to the \emph{Monotonic Translator} that predicts an intermediate monotonic MR $\bf z$. This is in turn fed to the \emph{Reorderer}, which outputs the final prediction $\bf y$.}
    \label{fig:concept}
\end{figure*}

We propose \textsc{TPol}, a two-step parsing approach with a modular framework made up of two components: a \emph{Monotonic Translator} and a \emph{Reorderer}. Figure \ref{fig:concept} shows how our semantic parser takes an input sentence $\bf x$ and predicts the corresponding MR $\bf y$. In the first step, $\bf x$ is fed to the Translator, which outputs a monotonic translation $\bf z$. In other words, $\bf z$ is the target MR that has been permuted so that it aligns monotonically to the input NL. Then, in a second step, $\bf z$ is fed to the Reorderer, which is trained to place the MR tokens back into the correct order to produce the final prediction $\bf y$.

The main idea behind \textsc{TPol} is decomposing the task into lexical translation and reordering, to learn more reliable translation patterns. We purport that modeling monotonic alignments eases the learning of novel pattern combinations of seen structures, improving compositional generalization.

An alternative approach would be to permute the NL inputs rather than the MRs monotonically. We do not follow this direction due to the observation that in semantic parsing, multiple NLs can map to the same MR. In other words, the NL domain is larger than that of the MRs, and thus we believe that learning to reorder the MRs is more feasible.


\subsection{Monotonic Translator}\label{lexical-translator}

The Monotonic Translator is responsible for making an initial prediction of the MR sequence, which will contain the correct tokens in monotonic order. To create the training bi-sequences, we use alignment information and permute the gold MR sequences to obtain a monotonic mapping with the NL. As a concrete example, consider the non-monotonic alignment in Figure \ref{fig:example-alignment-definition}a, and its monotonic translation in Figure \ref{fig:example-alignment-definition}b.

The translation task can be formulated in various ways. In our implementation, we work with two alternative approaches: a seq2seq Translator, and a tagger Translator. In the seq2seq formulation, $\bf x$ is fed into an encoder network, which produces a hidden vector. The hidden vector is fed to a decoder network which produces the output $\bf z$, i.e., the monotonically aligned MR. This can be implemented, for example, with a BART model \citep{lewis-etal-2020-bart}, which uses a bidirectional encoder and a left-to-right decoder. In our experiments, we use the multilingual version of BART \citep{liu-etal-2020-multilingual-denoising}. In the tagging formulation, the Translator assigns an MR token to each token in $\bf x$, obtaining the monotonic translation $\bf z$ by explicitly aligning in a token-by-token fashion. We implement this with a BERT model \citep{devlin-etal-2019-bert} and we use its classification head as the tagger. 

A crucial difference between the seq2seq and the tagger Translator is that the latter needs $\bf x$ and $\bf z$ to be the same length. The seq2seq Translator can learn to perform deletion operations from the raw NL, without needing epsilons in the input to perform insertions. By contrast, the tagger Translator needs insertions to be performed on $\bf x$ before predicting $\bf z$. In general, NL sequences are significantly longer than the MR sequences, i.e., most epsilons are in the MR sequence. In other words, deletions are more frequent than insertions. 

However, for some datasets, some alignments contain epsilons in the NL sequence: at prediction time, we will not know where insertions might occur, and thus we need a way to predict them. For this purpose, for every token followed by an epsilon in the train split, we add an epsilon after it at test time. We saw that this strategy was sufficient in our experiments. Alternatively, this step could be done by a trained model or with a rule-based system similar to \citet{ribeiro-etal-2018-local}.

\subsection{Reorderer}
The Reorderer module is responsible for taking the monotonic predictions of the Translator and putting them back into the correct order to obtain the final prediction. This model is trained from pairs of MR sequences $(\bf z, \bf y)$ where the input $\bf z$ is a monotonically permuted MR and the output $\bf y$ is the target MR in its correct order. These training pairs can be generated from the alignment annotations.

Similarly to the Translator module, the Reorderer can be implemented both as a seq2seq model and as a tagger. We use mBART in the seq2seq formulation and BERT as a tagger in our experiments. Note that we do not enforce the output to be a permutation of the input.

\section{Experiments}
\label{experiments}
\subsection{Datasets}
\label{datasets}
We test \textsc{TPol} on two semantic parsing datasets, training with gold and automatically generated alignments in multiple languages, on standard IID partitions and the more challenging compositional ones.

\subsubsection{\textsc{GeoAligned}}
\textsc{GeoAligned} \citep{locatelli-quattoni-2022-measuring} augments the popular \textsc{Geo} semantic parsing benchmark \citep{aaai-zelle-1996} with token alignment annotations. The dataset contains questions about US geography and corresponding meaning representation using the FunQL formalism \citep{kate-etal-2005-learning}. In total, there are 880 examples, all annotated with token alignments. We evaluate on three partitions: question (\textsc{?}), query (\textsc{Q}) and length (\textsc{LEN}). The question partition is a standard IID split where test and train are sampled from the same distribution. The query partition, introduced by \citet{finegan-dollak-etal-2018-improving}, is designed to be compositional by ensuring that the templates of the MRs in the test set are never seen during training. The length partition, introduced by \citet{herzig-berant-2021-span}, assigns the longest sequences to the test.

The dataset comes in English, Italian and German: in this way we can test our approach across different languages. In our experiments, we do not anonymize constants: in other words, we keep the original NL and MR sequences which include names of cities, states, etc. We follow \citet{wang-etal-2021-structured-nips} in removing brackets.
\subsubsection{\textsc{ScanSP}}
\textsc{ScanSP} \citep{herzig-berant-2021-span} is a set of navigational commands presented in natural language paired with action sequences. It is based on the \textsc{Scan} dataset by \citet{pmlr-v80-lake18a}, which does not contain program MRs. \citet{herzig-berant-2021-span} translated the sequences into programs to obtain a semantic parsing version of the dataset. Besides the IID split, we test on the compositional partitions based on the "right" (\textsc{RX}) and "around right" (\textsc{ARX}) primitives from \citet{loula-etal-2018-rearranging}. \textsc{ScanSP} has 20,910 commands distributed roughly as 12,000 train, 3,000 validation, and 4,000 test examples. 

The \textsc{ScanSP} dataset does not come with alignments. Therefore we employ the IBM models \citep{brown-etal-1993-mathematics} to generate them automatically using the GIZA++ toolkit \citep{giza++}. We also do this for \textsc{Geo} to compare the performance of \textsc{TPol} when trained with gold and automatic alignment annotations.

\begin{table*}[t]
\begin{center}
\begin{tabular}{lcccccccccccc}
    \toprule
    & \multicolumn{9}{c}{\textbf{\textsc{Geo}}} & \multicolumn{3}{c}{} 
    \\ \cmidrule(lr){2-10}
    \textbf{Model} & \multicolumn{3}{c}{\bf EN} & \multicolumn{3}{c}{\bf IT} & \multicolumn{3}{c}{\bf DE} & \multicolumn{3}{c}{\textbf{\textsc{ScanSP}}} 
    \\ \cmidrule(lr){2-4} \cmidrule(lr){5-7} \cmidrule(lr){8-10} \cmidrule(lr){11-13}
    & \textbf{?} & \textbf{Q} & \textbf{LEN} & \textbf{?} & \textbf{Q} & \textbf{LEN} & \textbf{?} & \textbf{Q} & \textbf{LEN} & \textbf{IID} & \textbf{RX} & \textbf{ARX}
    \\\cmidrule(lr){2-2} \cmidrule(lr){3-3} \cmidrule(lr){4-4} \cmidrule(lr){5-5} \cmidrule(lr){6-6} \cmidrule(lr){7-7} \cmidrule(lr){8-8} \cmidrule(lr){9-9} \cmidrule(lr){10-10} \cmidrule(lr){11-11} \cmidrule(lr){12-12} \cmidrule(lr){13-13}
    LSTM & 52.9 & 24.9 & 5.0 & 46.4 & 18.1 & 4.3 & 42.9 & 17.6 & 3.2 & 100 & 24.4 & 1.1\\
    mT5 & 80.0 & 60.0 & 19.3 & 73.2 & 44.9 & 20.4 & 68.2 & 47.8 & 18.6 & 100 & 41.2 & 99.8\\
    mBART & 87.5 & 69.4 & 27.5 & \textbf{86.6} & 76.4 & 23.3 & \textbf{75.5} & 56.3 & 18.2 & 100 & 99.4 & 100 \\
    \hline  
    \textsc{LeAR} & - & 84.1 & - & - & - & - & - & - & - & - & - & - \\
    \textsc{SpanBased} & \textbf{87.7} & 74.6 & \textbf{55.0} & - & - & - & - & - & - & 100 & 100 & 100 \\
    \multicolumn{1}{r}{- gold} & 66.4 & 51.8 & 24.6 & 49.6 & 37.3 & 10.4 & 40.4 & 21.4 & 5.0 & 100 & 100 & 100 \\
    \textsc{ReMoto} & 75.2 & 43.2 & 23.2 & - & - & - & 55.6 & 22.3 & 16.6 & 100 & - & - \\
    \hline 
    \textsc{TPol} & 87.3 & \textbf{87.8} & 41.9 & 85.9 & \textbf{81.6} & \textbf{31.3} & 73.3 & \textbf{69.4} & \textbf{22.9} & - & - & -\\
    \multicolumn{1}{r}{- gold} & 85.8 & 79.0 & 35.6 & 83.6 & 75.1 & 20.2 & 73.8 & 60.7 & 17.5 & 100 & 99.4 & 100 \\
    \bottomrule
\end{tabular}
\end{center}
\caption{Exact-match accuracy of all models on \textsc{GeoAligned} and \textsc{ScanSP} datasets. \textsc{?} stands for question, \textsc{Q} for query and \textsc{LEN} for the length partition. \textsc{RX} stands for right and \textsc{ARX} for around right partitions. \textsc{LeAR} and \textsc{ReMoto} both anonymize constants in \textsc{Geo}, and the results are taken directly from the respective papers.}
\label{tab:exact_match_main_res}
\end{table*}

\subsection{Models for comparisons}
We compare with competitive baselines and state-of-the-art models that do not leverage alignments and competing models that do.\\
 
\textbullet\ \textbf{LSTM}: a standard seq2seq model with a bi-directional LSTM encoder and an LSTM decoder with attention \citep{bahdanau-etal-2015-neural}. We use pre-trained GloVe embeddings for the three languages: English \citep{pennington2014glove}, Italian and German \citep{ferreira-etal-2016-jointly}.

\textbullet\ \textbf{mBART} \citep{liu-etal-2020-multilingual-denoising}: a multilingual version of \textsc{BART} \citep{lewis-etal-2020-bart}, a pre-trained Transformer-based seq2seq model that has been successfully applied to parsing \citep{bevilacqua-etal-2021-one}.

\textbullet\ \textbf{mT5} \citep{xue-etal-2021-mt5}: a multilingual version of T5 \citep{raffel-etal-2020-exploring}, pre-trained on the mC4 dataset \citep{xue-etal-2021-mt5}.

\textbullet\ \textsc{\textbf{SpanBased}} \citep{herzig-berant-2021-span}: a semantic parser that predicts a span tree over an input utterance trained with gold alignment trees. The authors provided annotations for the English version of \textsc{Geo} and \textsc{ScanSP}. For the other languages of \textsc{Geo} we train without gold alignments. We use their model without the lexicon, as that would be unfair with respect to the other models.

\textbullet\ \textsc{\textbf{LeAR}} \citep{liu-etal-2021-learning-algebraic}: a model that learns to recombine structures recursively by predicting a latent syntax tree and assigning semantic operations to non-terminal nodes. \textsc{LeAR} explicitly uses alignments using a phrase table.

\textbullet\ \textsc{\textbf{ReMoto}} \citep{wang-etal-2021-structured-nips}: a model that first reorders the tokens in the NL and then predicts the MR. \textsc{ReMoto} is not trained with gold alignments.\footnote{For \textsc{LeAR} and \textsc{ReMoto}, we report results directly from the respective papers, noting that in their setting constants, such as names of states, cities, and so on, are anonymized.}

\subsection{Evaluation metric}

We follow the standard practice of using exact-match accuracy for evaluation: the predicted MR is correct only if it is the same as the gold.

\subsection{Main Results}
\label{main_results}

We report the results of our experiments in Table \ref{tab:exact_match_main_res}. For \textsc{TPol}, the choice of the modules’ architecture is validated on the development set, and we report a performance breakdown in Section \ref{architectural-studies}.

We first consider the results of \textsc{TPol} trained with gold alignments. On the \textsc{Geo} dataset, the LSTM and mT5 achieve the lowest performance in all the partitions. Looking at the question partition (\textsc{?}), the models show similar performance to mBART and \textsc{SpanBased}, which is not surprising as the test split does not require compositional generalization. On the query partition (\textsc{Q}), designed to test for compositional generalization, \textsc{TPol} shows significant improvements over all the other models across all languages. In English it obtains 87.3\% outperforming mBART (69.4\%), \textsc{SpanBased} (74.6\%) and \textsc{LEAR} (84.1\%). In Italian and German, it obtains 81.6\% and 69.4\% respectively, while mBART 67.4\% and 56.3\%. On the length partition (\textsc{LEN}), \textsc{TPol} does better than all the baselines across all languages, except for \textsc{SpanBased}, which fares better on LEN(English). This is only the case, however, when gold alignments are provided.

Looking at the results obtained without gold alignments, \textsc{TPol} shows considerable improvements over \textsc{ReMoto} and \textsc{SpanBased}. In particular, it improves on the English query partition obtaining 79\% against 43.2\% and 51.8\%, respectively. Furthermore, the accuracy does not drop significantly compared to \textsc{TPol} trained with gold alignments. We tested using automatic alignments from IBM models 3, 4, and 5 and picked the best out of the three. In general, all lead \textsc{TPol} to achieve similar performance. 

Finally, looking at \textsc{ScanSP}, as expected, the models designed for compositional generalization achieve perfect performance on the dataset. What is surprising is that also mBART can do so, contrary to other deep models. With some internal testing, we have seen that this is not the case for English \textsc{BART}, as opposed to the multilingual version. We hypothesize that model size and pre-training might be a factor of success for mBART.

\section{Error Analysis}

 \begin{table*}[h]
\begin{center}
\begin{tabular}{lcccccc}
    \toprule
    & \multicolumn{6}{c}{\textbf{\textsc{Geo} EN}}\\ \cmidrule(lr){2-7} 
    \textbf{Model} & \multicolumn{2}{c}{\textbf{?}} & \multicolumn{2}{c}{\textbf{Q}} & \multicolumn{2}{c}{\textbf{LEN}}
    \\ \cmidrule(lr){2-3} \cmidrule(lr){4-5} \cmidrule(lr){6-7}
    & {\textbf{MN}} & {\textbf{NMN}} & {\textbf{MN}} & {\textbf{NMN}} & {\textbf{MN}} & {\textbf{NMN}}
    \\ \cmidrule(lr){2-2} \cmidrule(lr){3-3} \cmidrule(lr){4-4} \cmidrule(lr){5-5} \cmidrule(lr){6-6} \cmidrule(lr){7-7}
    mBART &  90.4 & 81.0 & 67.1 & 76.5 & 29.2 & 25.1 \\ 
    \textsc{SpanBased} & 94.7 & 73.7 & 89.6 & 39.8 & 68.1 & 35.9 \\
    \textsc{TPol} & 89.9 & 81.4 & 93.7 & 69.9 & 55.5 & 23.2 \\
    \bottomrule
\end{tabular}
\end{center}
\caption{Performance breakdown of \textsc{TPol} over monotonic (MN) and non-monotonic (NMN) sequences in \textsc{GeoAligned} English. In Appendix \ref{sec:appendix2} we report the number of MN and NMN examples.}
\label{tab:mn_breakdown}
\end{table*}

\begin{table}[t]
\begin{center}
\begin{tabular}{lccc}
    \toprule
    & \multicolumn{3}{c}{\textbf{\textsc{Geo} EN}}
    \\ \cmidrule(lr){2-4} 
    \textbf{Module} & {\textbf{?}} & {\textbf{Q}} & {\textbf{LEN}}
    \\ \cmidrule(lr){2-2} \cmidrule(lr){3-3} \cmidrule(lr){4-4} 
    Translator & 86.1 & 86.2 & 42.5 \\ 
    Reorderer  & 87.6 & 87.3 & 57.1 \\ 
    \bottomrule
\end{tabular}
\end{center}
\caption{Performance breakdown of \textsc{TPol} modules.}
\label{tab:breakdown_error}
\end{table}

\begin{table*}[h]
\begin{center}
\begin{tabular}{lccccccccc}
    \toprule
    & \multicolumn{9}{c}{\textbf{\textsc{Geo}}}
    \\ \cmidrule(lr){2-10}
    \textbf{Model} & \multicolumn{3}{c}{\bf EN} & \multicolumn{3}{c}{\bf IT} & \multicolumn{3}{c}{\bf DE} 
    \\ \cmidrule(lr){2-4} \cmidrule(lr){5-7} \cmidrule(lr){8-10} 
    & \textbf{?} & \textbf{Q} & \textbf{LEN} & \textbf{?} & \textbf{Q} & \textbf{LEN} & \textbf{?} & \textbf{Q} & \textbf{LEN}\\
    \cmidrule(lr){2-2} \cmidrule(lr){3-3} \cmidrule(lr){4-4} \cmidrule(lr){5-5} \cmidrule(lr){6-6} \cmidrule(lr){7-7} \cmidrule(lr){8-8} \cmidrule(lr){9-9} \cmidrule(lr){10-10}
    \textsc{Bert2Bert} & 74.9 & 84.6 & {\bf 41.9} & 74.0 & 74.2 & {\bf 31.3} & 62.5 & 67.5 & {\bf 22.9} \\
    \textsc{Bert2}m\textsc{Bart} & 82.1 & {\bf 87.8} & 36.4 & 76.9 & {\bf 81.6} & 27.5 & 63.2 & 68.9 & 20.0 \\
    \textsc{Bert2}m\textsc{Bart silver} & 82.5 & 87.0 &  34.6 & 76.6 & 81.1 & 27.0 & 65.4 & {\bf 69.4} & 19.9 \\
    m\textsc{Bart2Bert} & 74.5 & 70.7 & 24.5 & 77.7 & 76.1 & 24.2 & 65.7 & 56.1 & 17.5 \\
    m\textsc{Bart2}m\textsc{Bart} & {\bf 87.3} & 72.2 &  25.2 & 85.7 &  76.3 & 23.1 & {\bf 73.3} & 59.2 & 16.3 \\
    m\textsc{Bart2}m\textsc{Bart silver} &  86.4 & 71.5 & 24.9 & {\bf 85.9} & 75.8 & 20.0 & 73.2 & 59.5 & 17.0 \\
    \bottomrule
\end{tabular}
\end{center}
\caption{Performance breakdown of \textsc{TPol} for different module architectures on \textsc{GeoAligned}.}
\label{tab:architectures_comparison}
\end{table*}

Table \ref{tab:breakdown_error} shows a breakdown of performance of \textsc{TPol} on the English version of \textsc{GeoAligned}. The results indicate the exact-match accuracy achieved by the two modules: we can check whether the model struggles more with the translation or reordering step. To analyze the Translator's performance, we regard the monotonically aligned MRs as the gold truth. For the Reorderer, we provide it with the monotonically aligned MRs in input. In other words, the evaluation of the Reorderer assumes that the Translator makes a correct prediction. 

The performance of the two modules is fairly similar, and, by comparing these results with Table \ref{tab:exact_match_main_res}, we see that the accuracy of each component is not much higher than the overall accuracy, suggesting that neither component is hampering performance more than the other. The only exception seems to be in the length partition, where the Reorderer does considerably better than the Translator.

Table \ref{tab:mn_breakdown} shows the breakdown of the performance over monotonically and non-monotonically aligned MRs. We can observe that, compared to \textsc{SpanBased}, \textsc{TPol} generally presents smaller drops in performance over the non-monotonic sequences. For instance, in the question partition, \textsc{SpanBased} drops from 94.7\% over the monotonic examples to 73.7\% over the non-monotonic, while \textsc{TPol} drops from 89.9\% to 81.4\%. When we look at the query partition, we see that for both of these models, the drop is much higher than the one on the question partition: \textsc{SpanBased} goes from 89.6\% to 39.8\%, while \textsc{TPol} drops from 93.7\% to 69.9\%. In other words, both models struggle more with non-monotonic examples when compositional generalization is required. \textsc{TPol}, however, still performs significantly better.

Surprisingly, mBART shows the opposite trend on the query partition, with the non-monotonic accuracy being higher than the monotonic one. By contrast, most of \textsc{TPol}'s improved performance comes from better modeling of the monotonic sequences ($67.1\% \rightarrow 93.7\%$). \textsc{TPol}'s results suggest that regular patterns in the non-monotonic sequences can be learned. Its generalization problems can be attributed to the difficulty of learning the more challenging non-regular patterns in a small dataset. On the other hand, mBART appears to have the capacity to model these challenging reorderings better. Still, this comes at the cost of failing on the regular monotonic ones, leading to a lower performance overall.

Interestingly, the \textsc{SpanBased} approach shows a similar improvement on monotonic sequences compared to mBART ($67.1\% \rightarrow 89.6\%$). This suggests that exploiting lexico-logical alignments allows models to capture the simpler patterns that mBART fails to learn. Most of \textsc{TPol}'s gains over \textsc{SpanBased} come from better modeling the non-monotonic examples in the query partition ($39.8\% \rightarrow 69.9\%$). This shows that the two-step approach offers the best of both worlds: it can capture the simple monotonic patterns  while maintaining reasonable performance over the more complex alignments on which \textsc{SpanBased} fails.  

\begin{figure}[t]
    \centering
    \includegraphics[width=\linewidth]{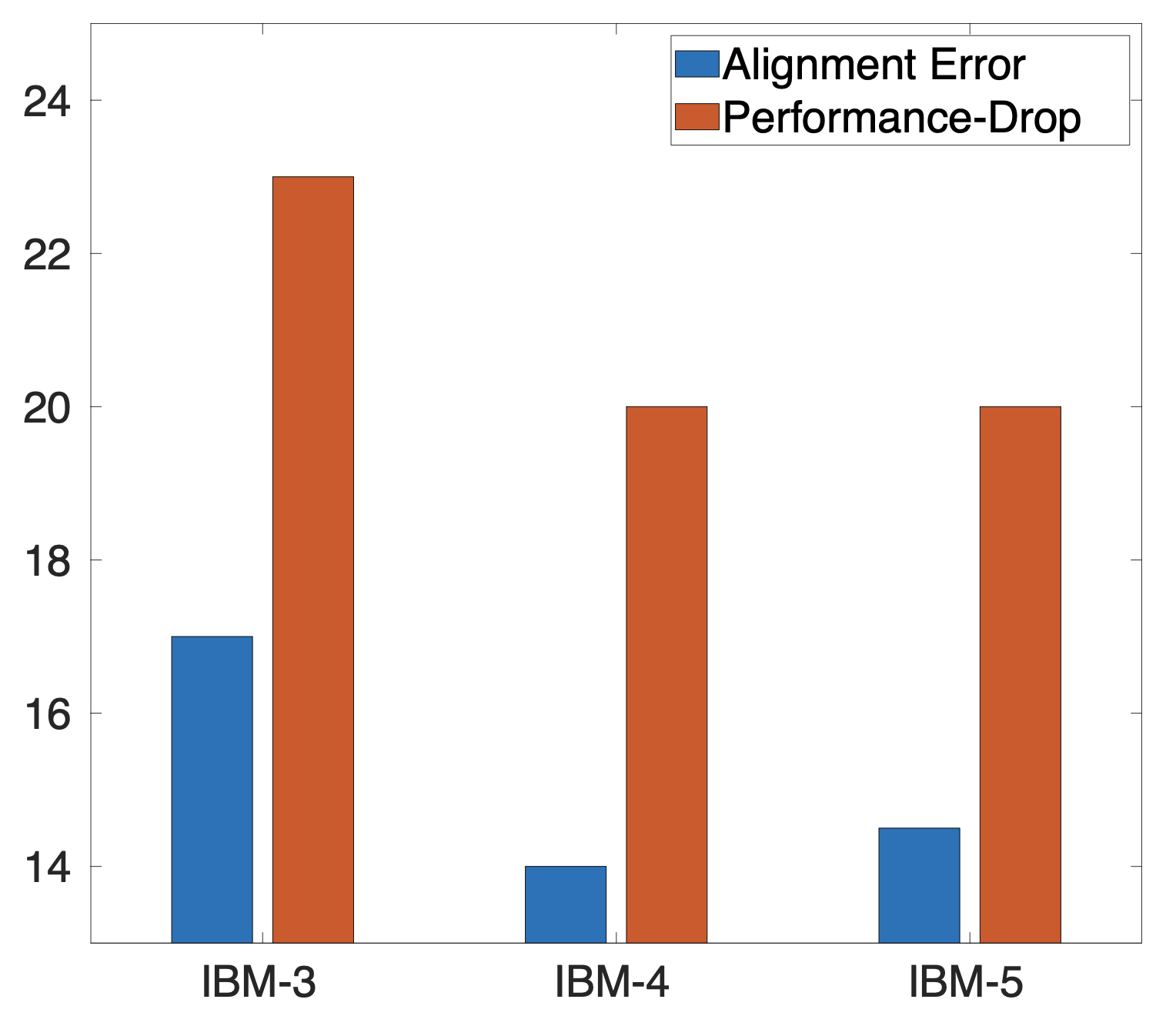}
    \caption{Average error of IBM alignment models over all partitions and languages on \textsc{GeoAligned}. We also plot the average drop in performance for each \textsc{TPol} model trained with IBM alignments with respect to the corresponding one trained with gold alignments.}
    \label{fig:giza_vs_per}
\end{figure}

Figure \ref{fig:giza_vs_per} shows the average drop in performance of the different models trained with automatic IBM alignments compared to the same model trained with gold alignments. We also report the corresponding alignment error, calculated as the percentage of bi-symbols that differ from the gold annotations from \textsc{GeoAligned}. We observe that, in general, higher alignment error is associated with a higher drop in performance. This validates the importance of the alignment information and points to improving the unsupervised alignment algorithm as a natural line of future work. We believe that one possible reason for the drop in performance when training with IBM alignments might be because the GEO dataset is of relatively small size, and the IBM models might have difficulty learning good alignments. That would explain why in contrast to GEO, the performance over SCAN is not affected by automatic alignments since SCAN is a much larger dataset.

\section{Architecture study}
\label{architectural-studies}

We emphasize here that our approach is an abstract, high-level methodology and does not place any constraint on the underlying architectures of the two components. We believe that different architectures, particularly specialized ones for each module, could be beneficial for parsing performance. We encourage further work to be carried out in this regard. To this purpose, we present some architectural studies using \textsc{BERT} \citep{devlin-etal-2019-bert} and mBART \citep{liu-etal-2020-multilingual-denoising} as components. We employ them as both Translator and Reorderer, examining all possible combinations. As explained in Section \ref{lexical-translator}, mBART is used as a standard seq2seq model, and \textsc{BERT} is employed with a classification head to function as a tagger for every input token. 

Additionally, when mBART is used as a Reorderer, we introduce a silver training setting. In the normal setting, the Reorderer is trained by taking the gold alignment annotations and outputting the meaning representation. In the silver setting, we use the predictions of the Translator model as training input. By doing so, the Reorderer trains on inputs that mimic more closely what it will actually receive at test time: this is done straightforwardly for a seq2seq model like mBART, while for our BERT tagger, every token in input needs to be aligned with a token in output, and when the input is corrupt it is not possible to achieve the same training technique.

In Table \ref{tab:architectures_comparison}, we present the results for our different architectural components. To distinguish among the different model combinations, we use a [Translator]2[Reorderer] naming convention, meaning that m\textsc{Bart2Bert} uses m\textsc{BART} as the Translator and \textsc{Bert} as the Reorderer. We observe that our two-step approach seems to be robust overall. 

We can discern trends in different architecture combinations, which can be helpful when choosing an architecture for a specific task. One important observation is that the architectures that use BERT as a Translator are consistently better than the ones using mBART over the compositional partitions. We hypothesize that the BERT Translator can achieve higher compositional generalization because it can better leverage alignment information to predict unseen combinations of observed training patterns. We believe this is because a tagger's predictions can be more naturally broken into parts that can be recombined. In contrast, encoder-decoder architectures fare better on the IID partition but struggle to generalize to unseen patterns. One possible reason is that these models have a harder time inducing local patterns that can be recombined since they encode and decode complete structures all at once.

\section{Conclusion}
\label{conclusion}
Seq2seq models have become increasingly popular in semantic parsing. However, they are limited in their abilities to generalize to unobserved structures. Here, we proposed \textsc{TPol}: a two-step parsing approach that leverages alignment annotations with a modular framework composed of a Translator and a Reorderer. 

We showed that \textsc{TPol} improves compositional generalization over conventional seq2seq models and over competing models that also leverage alignment information. Our results also showed that our approach is robust when trained with automatically generated alignments, demonstrating competitive results on two semantic parsing datasets. 

We have experimented with two possibilities for the Translator and Reorderer, but we believe that different architectural components could further improve performance. The divide-and-conquer strategy of breaking the problem into two simpler sub-tasks is designed to enable further component specialization. 

\section{Limitations}
Regarding the limitations of our approach, our experiments used the standard FunQL meaning representation. Transitioning to a different meaning representation might need some adaptation of the framework. In particular, the alignments between NL and MRs for other meaning representations might require more insertion and deletion operations. We might also expect that other MRs might require more reordering.

A second limitation of our work is training with gold alignments. We partially address this by training \textsc{TPol} with automatic alignments obtained with the IBM models. Still, we believe there is room for more work to be done so that this approach can be more easily scaled to datasets that do not have alignment annotations.

Despite \textsc{TPol}'s partial improvements on the length test splits, this type of partition remains challenging for all models. Here, models are required to generate predictions of greater length than what they have seen during training. This requires complex compositional productivity skills, i.e., recombining known constituents into larger structures. Further work is needed to address the limitation of the current state-of-the-art on compositional productivity benchmarks.

\section*{Acknowledgements}
We would like to thank the EACL area chair and the anonymous reviewers for their feedback, as well as the other members of the INTERACT group at the Universitat Politècnica de Catalunya for fruitful discussions on an earlier draft of this work. This work is supported by the European Research Council (ERC) under the European Union’s Horizon 2020 research and innovation program (grant agreement No.853459). This paper reflects the authors’ view only, and the funding agencies are not responsible for any use that may be made of the information it contains. The authors gratefully acknowledge the computer resources at Artemisa, funded by the European Union ERDF and Comunitat Valenciana, as well as the technical support provided by the Instituto de Fisica Corpus- cular, IFIC (CSIC-UV).

\bibliographystyle{acl_natbib}
\bibliography{anthology,custom}

\appendix

\section{Experimental details}
\label{sec:appendix}
For our experiments with \textsc{TPol} we report the average of three runs for every result. We select the hyperparameters with grid search on the development set performance stopping when there is no more improvement. We choose the learning rate among $1e^{-4}$, $1e^{-5}$ and $1e^{-6}$ and the batch size between the bounds of $4$ and $32$. Usually the best performing models choose a learning rate of $1e^{-5}$ and batch size of $8$. An experiment takes about $90$ minutes on a single Nvidia V100 GPU. Our \textsc{BERT} ($110$M parameters) and mBART ($680$M parameters) implementations are taken from the transformers library \citep{wolf-etal-2020-transformers}. We use for English \textit{bert-base-uncased}, for Italian \textit{dbmdz/bert-base-italian-uncased} and for German \textit{dbmdz/bert-base-german-uncased}. For mBART we use \textit{facebook/mbart-large-50}. For mT5 we use the \emph{google/mt5-small} pre-trained checkpoint from the Transformers library.

\section{\textsc{GeoAligned} statistics}
\label{sec:appendix2}
Table \ref{tab:stats} provides statistics from the English version of \textsc{GeoAligned} \citep{locatelli-quattoni-2022-measuring}. In particular, we report the number of examples that fall in the monotonic (MN) and non-monotonic (NMN) categories.
\begin{table}[h]
\begin{center}
\begin{tabular}{lccc}
    \toprule
    & \multicolumn{3}{c}{\textbf{\textsc{Geo} EN}}
    \\ \cmidrule(lr){2-4} 
    \textbf{Category} & {\textbf{?}} & {\textbf{Q}} & {\textbf{LEN}}
    \\ \cmidrule(lr){2-2} \cmidrule(lr){3-3} \cmidrule(lr){4-4} 
    \textbf{MN} & $194$ & $154$ & $162$ \\ 
    \textbf{NMN} & $86$ & $51$ & $118$ \\ 
    \bottomrule
\end{tabular}
\end{center}
\caption{Number of examples that belong to the monotonic (MN) and non-monotonic (NMN) categories in \textsc{GeoAligned} English.}
\label{tab:stats}
\end{table}
\end{document}